\documentclass{article}
\usepackage{spconf_,amsmath,graphicx}

\title{Towards Realizing the Value of Labeled Target Samples: a Two-Stage Approach for Semi-Supervised Domain Adaptation}
\name{Mengqun Jin$^{1}$, \quad Kai Li$^{2}$, \quad Shuyan Li$^{1}$, \quad Chunming He$^{1}$, \quad Xiu Li$^{1\ddagger}$}
  
  \address{$^{1}$Tsinghua Shenzhen International Graduate School, Tsinghua University, ShenZhen, China   \\
      $^{2}$ NEC Labs, America}
\begin{document}
%
\maketitle
\begin{abstract}


Semi-Supervised Domain Adaptation (SSDA) is a recently emerging research topic that extends from the widely-investigated Unsupervised Domain Adaptation (UDA) by further having a few target samples labeled, i.e., the model is trained with labeled source samples, unlabeled target samples as well as \textit{a few labeled} target samples.
Compared with UDA, the key to SSDA lies how to most effectively utilize the few labeled target samples. Existing SSDA approaches simply merge the few precious labeled target samples into vast labeled source samples or further align them, which dilutes the value of labeled target samples and thus still obtains a biased model. To remedy this, in this paper, we propose to decouple SSDA as an UDA problem and a semi-supervised learning problem where we first learn an UDA model using labeled source and unlabeled target samples and then adapt the learned UDA model in a semi-supervised way using labeled and unlabeled target samples. By utilizing the labeled source samples and target samples separately, the bias problem can be well mitigated. We further propose a consistency learning based mean teacher model to effectively adapt the learned UDA model using labeled and unlabeled target samples. Experiments show our approach outperforms existing methods.   

\end{abstract}

\section{Introduction}
\label{sec:intro}




Domain adaptation (DA) studies the performance degradation problem of a model trained on a source dataset when tested on an out-of-distribution target dataset. Most existing methods tackle this problem by learning an adaptive model using labeled source samples and unlabeled target samples. These methods are called Unsupervised (target samples are not labeled) Domain Adaptation (UDA) methods. Recently some works \cite{li2021cross,saito2019semi,singh2021clda,yang2021deep,wang2022cross} started investigating the effect of having a few target samples labeled, and formulated a new research topic called Semi-Supervised Domain Adaptation (SSDA), which aims to learn a better adaptive model using labeled source samples, unlabeled target samples, as well as \textit{a few} (e.g., one sample per class) labeled target samples. 

A naive way to leverage the few labeled target samples is to merge them into the labeled source samples and train a model in an UDA fashion \cite{saito2019semi,xie2020unsupervised}. However, due to the overwhelming dominance of the labeled source samples, the value of the few labeled target samples cannot be fully realized. Some methods strive to mitigate this problem: \cite{li2021ecacl} performed explicitly class-wise domain alignment using labeled samples from both domains, aiming to push close samples from the same class across domains. Similarly, \cite{singh2021clda} use contrastive learning to achieve class-wise alignment. \cite{li2021cross} instead used adversarial adaptive clustering \cite{li2021cross} to encourage samples from the same classes to agglomerate around the labeled target samples. Albeit achieving improved performance, these methods shall still suffer from the imbalance problem as the model could easily learn how to slightly adjust the classes boundaries to include several more labeled samples, resulting in a model still biased towards the source domain.   

\begin{figure}
\centering
\includegraphics[width=\linewidth]{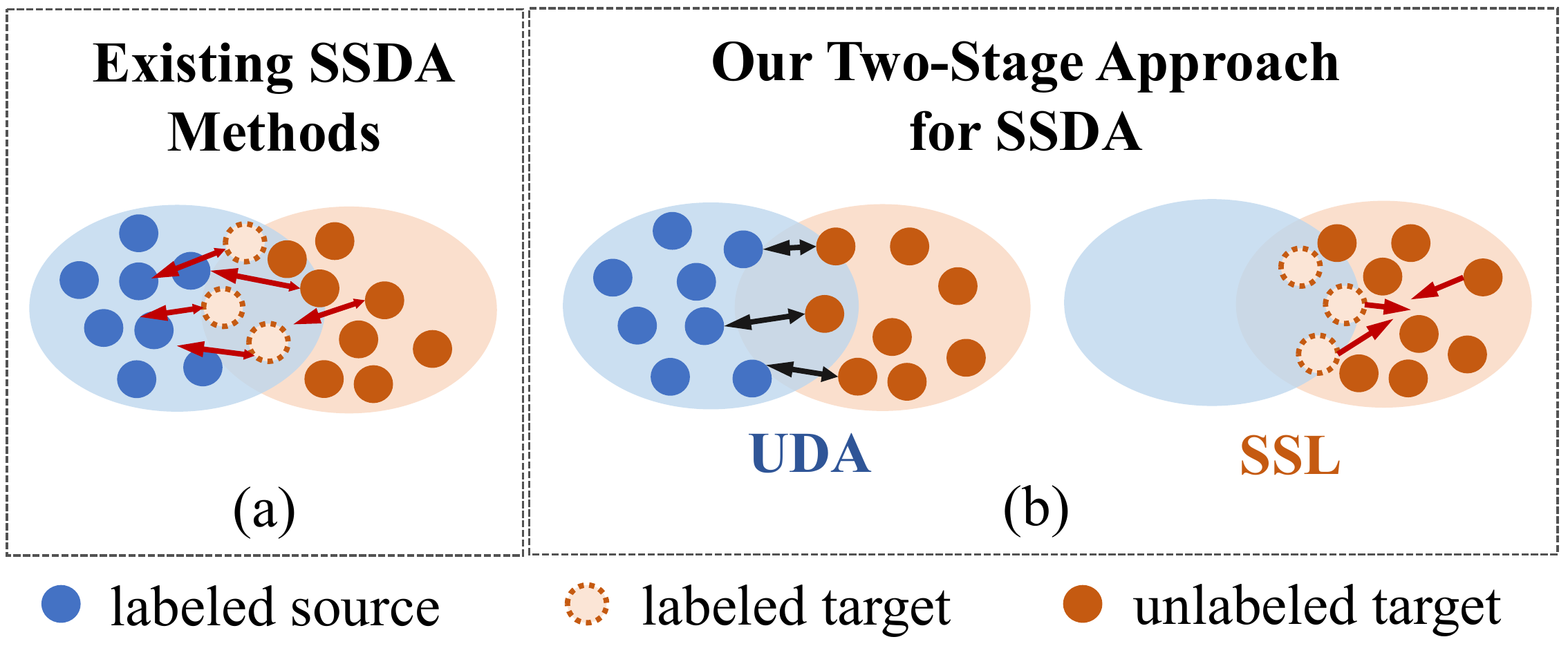}\vspace{-4mm}
\caption{Schematic difference of the proposed method from existing methods.
}
\label{UDA}
\vspace{-5mm}
\end{figure}

In this paper, we approach SSDA in a new perspective which fundamentally resolves the imbalance problem by utilizing labeled source samples and labeled target samples in two separated stages. In ths first stage, we use labeled source samples and unlabeled target samples to train an UDA model. Next, we aim to adapt this UDA method using only target samples, both labeled and unlabeled, in a Semi-Supervised Learning (SSL) fashion. 
As in the second stage, only the labeled target samples provide supervision signals, their values are more likely to be realized, without risking being diluted by vast label source samples, facilitating to produce a model more favorable to the target domain. Fig. \ref{UDA} shows the contrast of our method to existing SSDA methods. 

Decoupling SSDA as an UDA problem and an SSL problem brings another benefit: In the second stage, we no longer need access to the labeled source samples which could be under protection, privacy sensitive, or large-scale such that it is less likely to be shared for adaptation \cite{pmlr-v119-liang20a,yang2021exploiting}. Our two-stage solution would thus be favored in the scenario where solution provider only offers models which are trained with protected labeled source data and cheap unlabeled target data and the scenario with resource-contracted devices.

Another imbalance problem arises immediately with the proposed two-stage solution - the imbalance between labeled target samples and unlabeled target samples. We propose a novel Consistency Learning based Mean Teacher (CLMT) model to handle this. For each unlabeled target image, we generates one \textit{weakly-augmented} image and one \textit{strong-augmented} image. We use a teacher model to produce a pseudo label for the image and use the pseudo label to train the student model in supervised way by taking the strongly-augmented image as input. In this way, we gradually turn unlabeled target samples into labeled ones and thus alleviate this newly-arising imbalance problem.

\begin{figure}
\centering
\includegraphics[width=\linewidth]{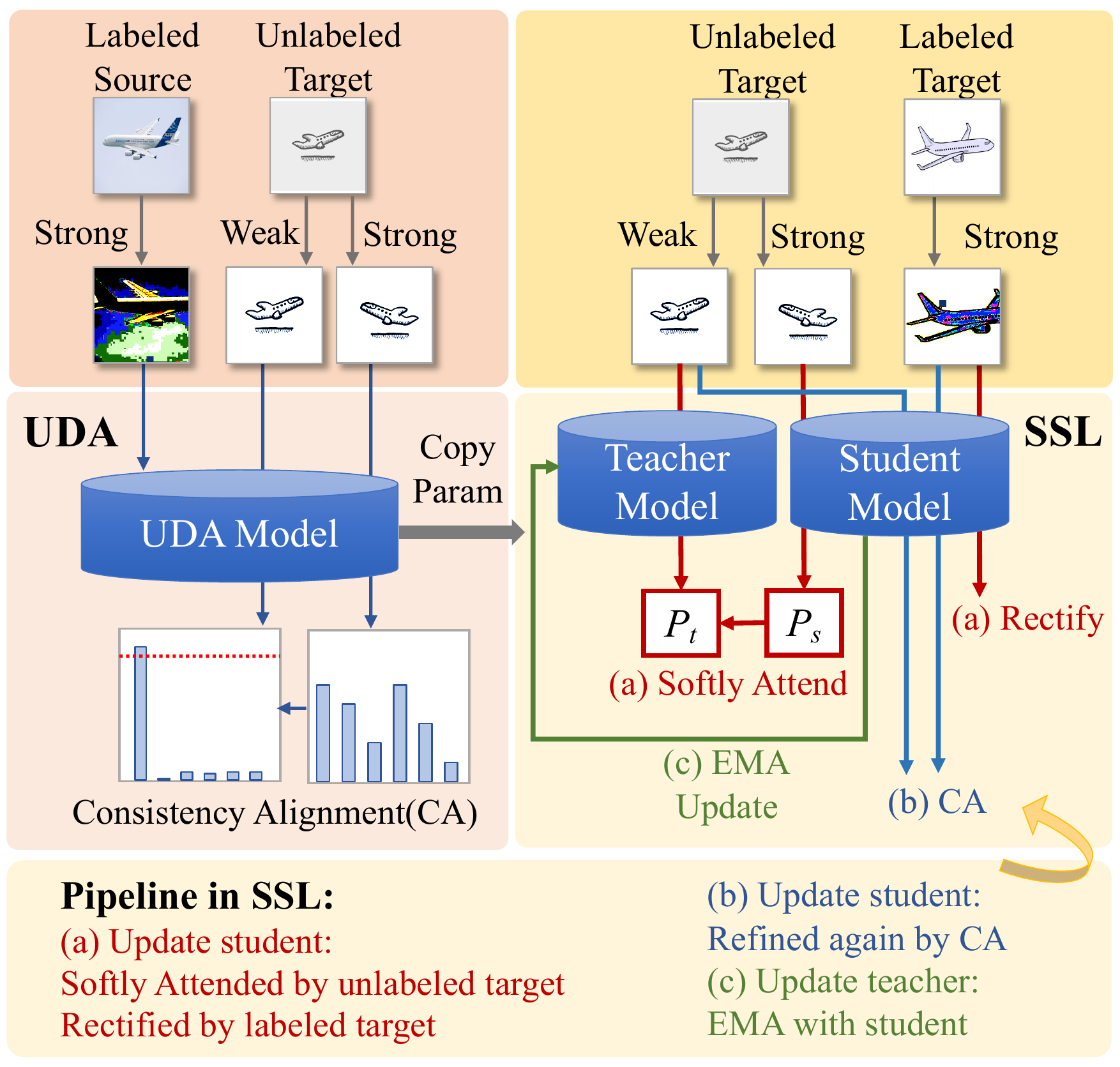}\vspace{-3mm}
\caption{Framework of our proposed method. 
}
\label{framework}
\vspace{-4mm}
\end{figure}

The contributions of this paper are as follows: (1) We propose to address the SSDA problem in a new perspective by decoupling it as a UDA problem and a SSL problem, which not only resolves the imbalance problem between labeled source samples and labeled target samples, but also offers a solution to protecting labeled source samples. (2) We propose the consistency learning based mean teacher model which is able to mitigate the imbalance problem between unlabeled and labeled target samples.
(3) Our method outperforms existing SSDA methods on the common evaluation datasets.

\section{Method}
\label{sec:method}

In the setting of SSDA, we have a large bunch of source data $ S=\{{(\textbf{s}_i, y_i^s)}\}_{i=1}^{N_s}$, abundant unlabeled target domain data
$T_u=\{{\textbf{u}_i}\}_{i=1}^{N_u}$
 and only a few labeled target data $T_l=\{{(\textbf{t}_i, y_i^t)}\}_{i=1}^{N_t}$. Here, the labeled target data is usually 1-shot or 3-shot, which is extremely scarce compared to ${N_u}$. We name the labeled target samples as ``Anchors''. We use $ T $ to denote the target domain data, where $ T = T_u \cup T_l$ and $ T_u \cap T_l = \emptyset$. $S$ and $T$ have different distribution but share the same label space $Y = \{1, 2, ..., C\}$. $C$ is the number of classes. 
We aim to learn a domain adaptive model which performs well on the target domain. Let the model be $\theta = f \circ g$, where $f$ extracts features and feeds them into the classifier $g$ that outputs the predictions. Our method includes two stage, the UDA stage and the SSL stage.

\subsection{Stage I: Unsupervised Domain Adaptation}
\label{ssec:subhead}
We train a UDA model using the following learning objective: 
\begin{equation}
\begin{aligned}
\mathcal{L}_{uda} =\mathcal{L}_s(S) + \alpha\mathcal{L}_u(S, T_u),
\end{aligned}
\end{equation}
where $\mathcal{L}_s(S)$ is the supervised loss with labeled source samples, and $\mathcal{L}_u(S, T_u)$ is the domain alignment loss with labeled source and unlabeled target samples. $\alpha$ is a hyper-paremeter. 

Recent studies demonstrate that data augmentation \cite{cubuk2019autoaugment, cubuk2020randaugment} can significantly improve the generalization of supervised learning, especially in image classification and object detection.
To extract robust feature, we process label-rich source data $S$ with RandAugment \cite{cubuk2020randaugment}, a random augmentation technique, including color, brightness, rotation, sharpness, etc., and output $ S'=\{{(\textbf{s}'_i, y_i^s)}\}_{i=1}^{N_s}$.
\begin{equation}
\begin{aligned}
\mathcal{L}_s = \frac{1}{N_s}  \sum_ {(\textbf{s}'_i, y_i^s){\sim}S'}L(\theta(\textbf{s}'_i), y_i^s)
\end{aligned}
\end{equation}
where $L(p, y)$ is the cross-entropy loss over labeled samples. 


For each unlabeled target sample, we apply both weak augmentations $\varphi$
and strong augmentation $\psi$.
\begin{equation}
\begin{aligned}
\textbf{u}_i^w = \varphi(\textbf{u}_i), \hspace{10pt}
\textbf{u}_i^s = \psi(\textbf{u}_i).
\end{aligned}
\end{equation}

We feed $\textbf{u}_i^w$ and $\textbf{u}_i^s$ into the model $\theta$ and output the predictions $\textbf{p}_i^w$ and $\textbf{p}_i^s$. If the max prediction $\max({\textbf{p}_i^w})$ is greater than the threshold $\mu$, it indicates that the prediction has strong discrimination and tendency, thus regarded as a pseudo label $\tilde{\textbf{p}}^w = \mathrm{argmax}({\textbf{p}}^w)$, which returns a one-hot label to supervise the strong augmented unlabeled prediction $\textbf{p}_i^s$.

\begin{equation}
\begin{aligned}
\mathcal{L}_u = \sum_{\textbf{u}_i {\sim}U} [\mathbbm{1}(\max({\textbf{p}}^w) \geq \mu) H(\tilde{\textbf{p}}^w , {\textbf{p}}^s)],
\end{aligned}
\end{equation}
$H(. , .)$ is the cross-entropy of two possibility distributions. $\mathbbm{1}(.)$ is an indicator function that filters and returns the satisfied $\textbf{p}_i^w$ statistically.

\begin{figure}
\centering
\includegraphics[scale=0.3]{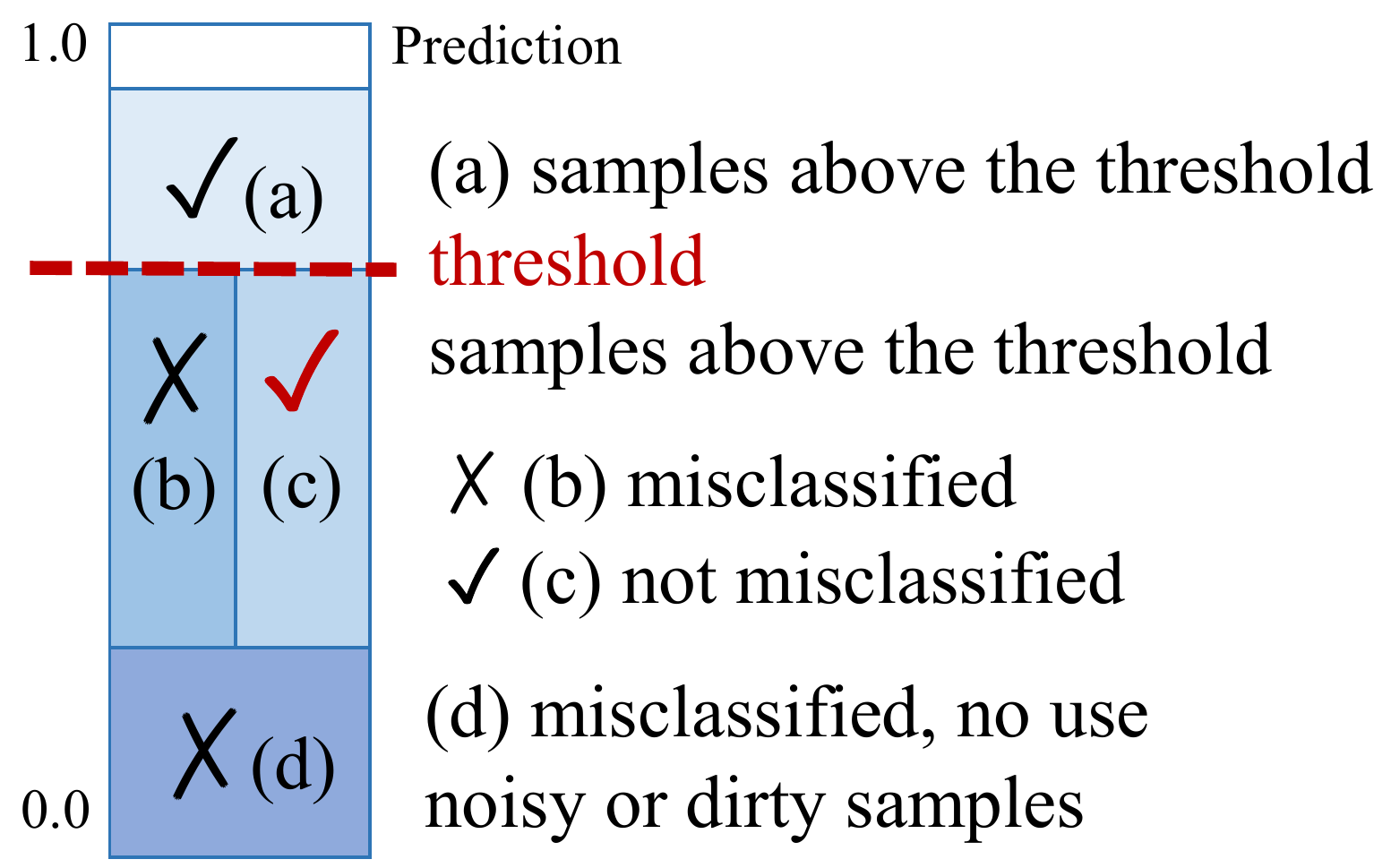}\vspace{-3mm}
\caption{CLMT rescues samples(c) whose $\max({\textbf{p}^w})$ below the threshold but not misclassified.}
\label{threshold}
\vspace{-4mm}
\end{figure}

\subsection{Stage II: Semi-Supervised Learning}
\label{ssec:subhead}
\vspace{-1mm}
After obtaining the UDA model, we start to adapt it to the target domain with proposed consistency learning based mean teacher (CLMT) model, by taking labeled and unlabeled target samples as input. 
Following existing SSL methods, we optimze our model with the following learning objective:
\vspace{-2mm}
\begin{equation}
\begin{aligned}
\mathcal{L}_{ssl} = \gamma \mathcal{L}_t(T_l) +\eta \mathcal{L}_d(U)
\end{aligned}
\vspace{-2mm}
\end{equation}
where $\mathcal{L}_t(T_l)$ is the supervised loss with labeled target samples $T_l$, and $\mathcal{L}_d(U)$ is the unsupervised loss with unlabeled target samples $U$.

To fully utilize the discriminative ``Anchors'', we extract strong augmented features $T'_l=\{{(\textbf{t}'_i, y_i^t)}\}_{i=1}^{N_t}$ and obtain explicit decision boundaries using $\mathcal{L}_t$. The soft distillation loss $ \mathcal{L}_{d} $ and target supervised loss $\mathcal{L}_{t}$ together optimize the student model $\theta_s$ shown in Figure \ref{framework}(a).
\vspace{-2mm}
\begin{equation}
\begin{aligned}
\mathcal{L}_t = \frac{1}{N_t}  \sum_ {(\textbf{t}'_i, y_i^t){\sim}T_l}L(\theta(\textbf{t}'_i), y_i^t)
\end{aligned}
\vspace{-2mm}
\end{equation}


In most UDA and SSDA methods \cite{shen2022connect,ganin2015unsupervised,wang2022cross,saito2019semi},
as the source domain provides sufficient supervised information, the good model drives the unlabeled target samples to hurdle $\max({\textbf{p}_i^w})$ above the threshold, the latter ones then join the training as supervisors. However, as shown in Figure \ref{threshold}, a bunch of correctly classified samples are ignored due to the threshold, which is expected to rescue.  Inspired by SimCLR v2 \cite{chen2020big}, we copy the UDA model $\theta$ as teacher model $\theta_t$ and student model $\theta_s$. We feed the strong augmented view $\textbf{u}^s$ into student model and the weak augmented view $\textbf{u}^w$ into the teacher model, and output the prediction $\textbf{p}^s$ and $\textbf{p}^w$, respectively. We assume that the teacher computes relatively precise soft labels for the student model to learn from. 
\begin{equation}
\begin{aligned}
\mathcal{L}_d = - \sum_{{u}_i {\sim}U} \theta_t(\textbf{u}^w) \log {\theta_s(\textbf{u}^s)} = - \sum  \textbf{p}^w\log{\textbf{p}^s}
\end{aligned}
\end{equation}


\vspace{-2mm}
After one round of rectification, we assume that the $\theta_s$ has preciser decision boundaries and can stride across more reliable unlabeled samples. We align the target domain implicitly using pseudo labeling and consistency alignment $\mathcal{L}_u$ again, shown in Figure \ref{framework}(b).


The teacher model $\theta_t$ is then updated by the EMA method, shown in Figure \ref{framework}(c). $\theta_{t-1} $ denotes the previous teacher model and $\theta_{s-1}$ represents the previous student model. $\theta_t$ indicates the updated teacher model. $\sigma$ is the EMA hyper-parameter for the teacher model to adapt.
\begin{equation}
\begin{aligned}
\theta_t = \sigma \theta_{t-1} + (1-\sigma)\theta_{s-1}
\end{aligned}
\end{equation}
After this overall pipeline, the student model is barely affected by the source domain. Thanks to the ``Anchor" and unlabeled target samples, the model is biased toward the target domain.

 \vspace{-2mm}
\section{Experiments}
\label{sec:experiment}

\vspace{-3mm}
\subsection{Setups}
\label{sssec:subsubhead}
\vspace{-2mm}
\textbf{Datasets.} We evaluate the effectiveness of our proposed approach on the latest domain adaptation dataset DomainNet \cite{peng2019moment}. It is a dataset of common objects, 6 different domains with 126 categories. Similar to MME \cite{saito2019semi}, we select 4 domains, Real(R), Clipart(C), Painting(P), and Sketch(S). We perform experiments and evaluate classification accuracy on 7 pairs, R→C, R→P, P→C, C→S, S→P, R→S, and P→R. For each pair, we evaluate 1-shot SSDA and 3-shot SSDA settings. The labeled target samples are randomly selected. 

\noindent\textbf{Implementation details.}
Similar to previous SSDA work, we choose Alexnet as our backbone networks and the MME method \cite{saito2019semi} as our baseline.
For fair comparisons, experimental settings in our proposed method, i.e, feature extractor, the linear classification layer, optimizer, learning rate, is set the same as MME. We train for 50K iterations. Threshold is fixed $\mu=0.95$. The hyper-parameters of different losses are $\alpha=1.0$, $\gamma=0.2$, $\eta = 0.8$. $\sigma$ is set as $0.99$.

\noindent\textbf{Compared methods.}
We compare with the following methods including
DANN,
ADR\cite{saito2017adversarial}, CDAN\cite{long2018conditional}, FAN\cite{kim2020attract}, BiAT\cite{jiang2020bidirectional}, MME\cite{saito2019semi}, Meta-MME\cite{li2020online}, PAC\cite{mishra2021surprisingly}, CLDA\cite{singh2021clda}, CDAC\cite{li2021cross}, and ECACL \cite{li2021ecacl}. 

\begin{table*}[t]
\small
\renewcommand{\tabcolsep}{4pt}
\begin{center}
 \caption{Results on the \textit{DomainNet} dataset on AlexNet. Best results are in \textbf{bold}.
 }
 \vspace{2mm}
\begin{tabular}{|l|cc|cc|cc|cc|cc|cc|cc|cc|c|} \hline

&\multicolumn{2}{|c|}{R$\rightarrow$C}
&\multicolumn{2}{|c|}{R$\rightarrow$P}
& \multicolumn{2}{|c|}{P$\rightarrow$C}
& \multicolumn{2}{|c|}{C$\rightarrow$S}
& \multicolumn{2}{|c|}{S$\rightarrow$P}
& \multicolumn{2}{|c|}{R$\rightarrow$S}
& \multicolumn{2}{|c|}{P$\rightarrow$R}
&\multicolumn{2}{|c|}{Mean} \\ \cline{2-17}

&1\scriptsize{-shot}&3\scriptsize{-shot}
&1\scriptsize{-shot}&3\scriptsize{-shot}
&1\scriptsize{-shot}&3\scriptsize{-shot}
&1\scriptsize{-shot}&3\scriptsize{-shot}
&1\scriptsize{-shot}&3\scriptsize{-shot}
&1\scriptsize{-shot}&3\scriptsize{-shot}
&1\scriptsize{-shot}&3\scriptsize{-shot}
&1\scriptsize{-shot}&3\scriptsize{-shot}  \\ \hline

ST  & 43.3      & 47.1 & 42.4   & 45.0 & 40.1   & 44.9 & 33.6   & 36.4 & 35.7   & 38.4 & 29.1 & 33.3 & 55.8   & 58.7 & 40.0 & 43.4 \\
 DANN    & 43.3      & 46.1 & 41.6   & 43.8 & 39.1   & 41.0 & 35.9   & 36.5 &36.9   & 38.9 & 32.5 & 33.4 & 53.6   & 57.3 & 40.4 & 42.4 \\
 ADR     & 43.1      & 46.2 &    41.4    & 44.4 &    39.3    & 43.6 & 32.8        &   36.4   &  33.1      &  38.9    &   29.1   &  32.4    & 55.9  & 57.3 & 39.2  & 42.7 \\
 CDAN    & 46.3      & 46.8 & 45.7   & 45.0 & 38.3   & 42.3 & 27.5   & 29.5 & 30.2   & 33.7 & 28.8 & 31.3 & 56.7   & 58.7 & 39.1 & 41.0 \\
 ENT   & 37.0      & 45.5 & 35.6   & 42.6 & 26.8   & 40.4 & 18.9   & 31.1 & 15.1   & 29.6 & 18.0 & 29.6 & 52.2   & 60.0 & 29.1 & 39.8 \\
 MME    & 48.9      & 55.6 & 48.0   & 49.0 & 46.7   & 51.7 & 36.3   & 39.4 & 39.4   & 43.0 & 33.3 & 37.9 & 56.8   & 60.7 & 44.2 & 48.2 \\
 Meta-MME  & -  & 56.4  & -  & 50.2  & -  & 51.9  & -   & 39.6  & -  & 43.7 & - & 38.7  & -  & 60.7  & -  & 48.7 \\
BiAT  & 54.2 & 58.6 & 49.2 & 50.6 & 44.0 & 52.0 & 37.7 & 41.9 & 39.6 & 42.1 & 37.2 & 42.0 & 56.9 & 58.8 & 45.5 & 49.4 \\
FAN     & 47.7  & 54.6 & 49.0 & 50.5 & 46.9 & 52.1 & 38.5 & 42.6 & 38.5 & 42.2 & 33.8 & 38.7 & 57.5 & 61.4 & 44.6 & 48.9 \\
PAC & 55.4 & 61.7 & 54.6 & 56.9 & 47.0 & 59.8 & 46.9 & \bf{52.9} & 38.6 & 43.9 & 38.7 & 48.2 & 56.7 & 59.7 & 48.3 & 54.7\\
CLDA &56.3 &59.9 &\bf{56.0} &57.2 &50.8 &54.6 &42.5 &47.3 &46.8 &51.4 &38.0 &42.7 &\bf{64.4} &67.0 &50.7 &54.3\\
CDAC  & 56.9 & 61.4 & 55.9 & 57.5 & 51.6 & 58.9 & 44.8 & 50.7 & 48.1 & \bf{51.7} & 44.1 & 46.7 & 63.8 & 66.8 & 52.1 &56.2 \\
ECACL  &55.8  &62.6  &54.0  &\bf{59.0}  &\bf{56.1}  &\bf{60.5}  &46.1 &50.6  &\bf{54.6}  & 50.3 & 45.0 & 48.4 & 62.3 & 67.4 & 52.8  &57.6\\
\hline
CLMT  &\bf{58.0} & \bf{63.1} & 55.7 & 57.7 & 55.6 & 60.3 & \bf{46.9} & 52.3 & 50.7 & 51.6 & \bf{49.3} & \bf{52.1} & 63.9 & \bf{68.2} &\bf{54.3} & \bf{57.8}

\\\hline
\end{tabular}
\label{result_domain_net}
\end{center}
\vspace{-6mm}
\end{table*}
 
\begin{table}[h]
\vspace{-8pt}
\small
\begin{center}
\caption{Comparison with decomposition in SSDA.}
\vspace{1mm}
\renewcommand{\tabcolsep}{7pt}
\begin{tabular}{c|c|c}\hline
Method &1-shot &3-shot \\\hline
Co-training SSDA & 45.0 & 48.4\\
Single Stage1 UDA  & 43.3  & 43.3\\
Single Stage2 SSL &24.2 &26.3 \\
\textbf{CLMT}  & 49.3  & 52.1
\\\hline
\end{tabular}
\label{table_ablation_final0}
\end{center}
\vspace{-0.6cm}
\end{table}


\begin{table}[h]
\vspace{-8pt}
\small
\begin{center}
\caption{Ablation study of the UDA stage in 3-shot.}
\vspace{1mm}
\renewcommand{\tabcolsep}{7pt}
\begin{tabular}{c|c|c}\hline
Method &AlexNet &ResNet34 \\\hline
Source & 30.5 &45.1\\
+Anchors & 33.3 & 50.1\\
+Anchors+Aug  & 37.6  & 54.8\\
+FixMatch &45.6 &67.2
\\\hline
\end{tabular}
\label{table_ablation_final1}
\end{center}
\vspace{-0.7cm}
\end{table}


\begin{table}[h]
\vspace{-9pt}
\small
\begin{center}
\caption{Ablation study of the SSL stage.}
\vspace{1mm}
\renewcommand{\tabcolsep}{7pt}
\begin{tabular}{c|c|c}\hline
Method &1-shot &3-shot \\\hline
UDA  & 43.3  & 43.3
\\
+MT & 44.8 & 44.9
\\
+FixMatch & 44.7  & 44.7
\\
+Anchors & 47.1 & 49.0
\\
+Anchors+FixMatch & 47.1 & 49.3
\\
+Anchors+MT & 47.4  & 49.1
\\
\hline
\textbf{CLMT}  & 49.3  & 52.1
\\\hline
\end{tabular}
\label{table_ablation_final2}
\end{center}
\vspace{-0.8cm}
\end{table}

\vspace{-6mm}
\subsection{Comparative Results}
\label{sssec:subsubhead}
\vspace{-1mm}
Comparing with the existing SSDA methods, we can see from the Table~\ref{result_domain_net} that our method attains 10.2(1-shot) and 9.6(3-shot) point gains over the baseline MME and surpasses the existing best performing three approaches, i.e., CLDA, CDAC and ECACL-P. CLMT achieves the best accuracy in 6 cases, including R→C(1$\&$3-shot), C→S(1-shot), R→S(1$\&$3-shot), P→R(3-shot), shown below. Our method with 6 best pairs significantly exceeds other methods, e.g. R→S(+4.3, +3.7). In contrast, the most competitive method ECACL achieves the best accuracy in only 4 cases, including R→P(3-shot), P→C(1$\&$3-shot), S→P(1-shot), shown below. For the rest 8 cases, we still achieve competitive results.

An interesting observation from Table~\ref{result_domain_net} is that, for different pairs, the gain varies within [0.9, 5.4] when adding 2-shot per-class. We suspect that a tug-of-war happens on the feature extractor to grab the data distribution between the two domains. The learning difficulty and dragging ability are diverse for each domain. The larger amount or better ability, the better performance as a target domain, etc Real. {$\frac{N_{T_u}}{N_{T_l}}$}varies between 60:1 and 200:1 and reveal that `Anchors' are tiny, comparing to the enormous($N_{T_u}$) unlabeled target samples. It is reasonable to leave the `Anchors' for the second stage, which amplifies the contribution and helps pin down the decision boundary.


 \vspace{-4mm}
\subsection{Analysis}
\label{sssec:subsubhead}
\vspace{-1mm}
We conduct all analysis with the adaptation experiment from the $\textit{Real}$ dataset to the $\textit{Sketch}$ dataset.

\noindent\textbf{Comparison with decomposition in SSDA.} 
To verify the effectiveness of decomposition, we compare our method with ECACL \cite{li2021ecacl}, shown in Table \ref{table_ablation_final0} as ``Co-training SSDA'', which includes data augment and consistency alignment, and prototype alignment. ``Single Stage1 UDA'' means UDA training without ``+Anchors''. ``Single Stage2 SSL'' applies only FixMatch\cite{sohn2020fixmatch}. The comparison shows that our decomposed two-stage approach outperforms the co-training ones, over 4.3$\%$(1-shot) and 3.7$\%$(3-shot). From ``Single Stage1 UDA'' to CLMT, 6.0$\%$ and 8.8$\%$ are gained, reconfirming that UDA model still has domain bias and room for improvement.

\noindent\textbf{Comparison with components in UDA.} 
We perform ablation studies on UDA under the 3-shot setting, to analyze the effectiveness of each loss term, including $L_{s}$ and $L_{u}$. As shown in Table \ref{table_ablation_final1}, ``Source'' only uses $L_{s}$, serving as baseline. ``Source'' and ``+Anchors'' have no augmentation, while ``+Anchors+Aug'' includes augmentation. It shows that despite labels are vacant in the target domain, the consistency alignment method is still fruitful for unlabeled target samples to pre-train a UDA model.

\noindent\textbf{Comparison with semi-supervised methods.}
As shown in Table~\ref{table_ablation_final2}, “+Anchors” indicates that the pretrained UDA model is then fine-tuned only on 1-shot or 3-shot labeled target samples using $L_t$. “+MT” and “+FixMatch” indicates $L_d$ Mean Teacher and $L_u$ FixMatch. “+Anchors+MT” and “+Anchors+FixMatch” are combinations of fine-tuning and semi-supervised methods. CLMT exceeds other semi supervised methods in solving domain adaptation challenge.

\vspace{-0.3cm}
\section{Conclusion}
\label{sec:conclusion}
\vspace{-3mm}
In this paper, we solve domain gap in a task decomposition way. We propose a simple but effective method for SSDA which decomposes SSDA into a two-stage approach, i.e., UDA and SSL. We well-pretrain a UDA model in the first stage. In the second stage, the target samples rectify and shift the model further to the target domain, using our method specially designed for the adaptation challenge. We outperforms existing methods on DomainNet, which demonstrates that the method performs well and alleviates domain bias.

 \vspace{-3mm}
\section{Acknowledgement}
 \vspace{-3mm}

This work was supported by Shenzhen Stable Supporting Program (WDZC202
00820200655001), the National Key R\&D Program of China 505 (Grant No.2020AAA0108303), and by Shenzhen Science and Technology Project (Grant No.JCYJ20200109143
041798) and Key Laboratory of next generation interactive media innovative technology (Grant No.ZDSYS20210623092001004).

\vfill\pagebreak

\bibliographystyle{IEEEbib}
\bibliography{refer}

\end{document}